\documentclass{article}
\usepackage{spconf,amsmath,graphicx,hyperref}

\usepackage{graphicx}
\usepackage{textcomp}
\usepackage{xcolor}
\def\BibTeX{{\rm B\kern-.05em{\sc i\kern-.025em b}\kern-.08em
    T\kern-.1667em\lower.7ex\hbox{E}\kern-.125emX}}

\usepackage{amsmath,amsthm,amssymb,mathtools,graphicx,enumitem,hyphenat,float,threeparttable}
\usepackage{amsfonts}
\usepackage{algorithm}
\usepackage{algpseudocode}
\usepackage{tabularx}
\usepackage{xspace}

\usepackage{multirow}
\usepackage{booktabs}
\usepackage{subfig}

\usepackage[most]{tcolorbox} 
\newtcbox{\mybox}[1][]{nobeforeafter,tcbox raise base,colframe=green!50!black,colback=green!10!white,top=0pt,bottom=0pt,left=0pt,right=0pt,before upper=\strut,#1}

\newcommand{\algnamelong}{\textbf{C}ompressed \textbf{A}ggregate \textbf{Fe}edback (\texttt{CAFe})\xspace}
\newcommand{\algname}{\texttt{CAFe}\xspace}

\newcommand{\norm}[1]{\left\lVert#1\right\rVert}
\newcommand{\sqn}[1]{{\left\lVert#1\right\rVert}^2}
\newcommand\ec[2][]{\ensuremath{\mathbb{E}_{#1} \left[#2\right]}}

\newcommand\ev[1]{\left \langle{#1}\right \rangle}

\newcommand\br[1]{\left({#1}\right)}

\newcommand{\R}{\mathbb{R}}
\newcommand{\compress}[1]{\mathcal{C}\br{#1}}

\usepackage{thmtools}

\declaretheorem[name=Lemma]{lemma}
\declaretheorem[name=Definition, sibling=lemma]{definition}

\declaretheorem[name=Assumption, style=plain, refname={assumption,assumptions}, Refname={Assumption,Assumptions}]{assumption}

\declaretheorem[name=Example]{example}

\declaretheorem[name=Corollary]{corollary}

\usepackage{hyperref,cleveref}
\crefname{assumption}{Assumption}{Assumptions}
\crefname{figure}{Fig.}{Figs.}
\crefname{equation}{Eq.}{Eqs.}

\title{Communication Compression for Distributed Learning without Control Variates}

\name{Tomas Ortega$^\star$,
Chun-Yin Huang$^\dagger$,
Xiaoxiao Li$^\dagger$
and
Hamid Jafarkhani$^\star$
\thanks{This work was supported in part by the NSF Award ECCS-2207457.
Tomas Ortega and Hamid Jafarkhani are with the Center for Pervasive Communications and Computing, University of California, Irvine, CA, USA 92697 (e-mail: \{tomaso, hamidj\}@uci.edu).
Chun-Yin Huang and Xiaoxiao Li are with the Computer Engineering Department at the University of British Columbia, Vancouver, BC, Canada, and Vector Institute, Toronto, ON, Canada (e-mail: \{chunyinh, xiaoxiao.li\}@ece.ubc.ca).}
}

\address{$^\star$ University of California, Irvine, CA, USA\\
$^\dagger$University of British Columbia, Vancouver, BC, Canada}

\begin{document}

\maketitle

\begin{abstract}
    Distributed learning algorithms, such as the ones employed in Federated Learning (FL), require communication compression to reduce the cost of client uploads.
    The compression methods used in practice are often biased, making error feedback necessary both to achieve convergence under aggressive compression and to provide theoretical convergence guarantees.
    However, error feedback requires client-specific control variates, creating two key challenges: it violates privacy-preserving principles and demands stateful clients.
    In this paper, we propose \textit{\algnamelong}, a novel distributed learning framework that allows highly compressible client updates by exploiting past aggregated updates, and does not require control variates.
    We consider Distributed Gradient Descent (DGD) as a representative algorithm and analytically prove \algname's superiority to Distributed Compressed Gradient Descent (DCGD) with biased compression in the non-convex regime with bounded gradient dissimilarity.
    Experimental results confirm that \algname{} outperforms existing distributed learning compression schemes.
\end{abstract}

\begin{keywords}
    Distributed Learning, Optimization, Federated Learning, Compression, Error Feedback.
\end{keywords}

\section{Introduction}\label{sec:intro}
In distributed learning, a central server coordinates the training of a global model using data stored across multiple clients.
The general problem formulation is to minimize the sum of client loss functions, which are typically non-convex.
We denote the global model as $x \in \R^d$, the client loss functions as $f_n : \R^d \to \R$, and the global loss function as
\begin{align}
    f(x) = \frac{1}{N} \sum f_n(x),\label{eq:global_loss}
\end{align}
where $N$ is the number of clients.
This formulation is prevalent in Federated Learning (FL)~\cite{communication_efficient}, a distributed learning paradigm designed for privacy preservation, where clients train the global model on their local data and send updates to the server for aggregation.
One of the main challenges in distributed learning is the communication cost associated with transmitting model updates from clients to the central server~\cite{advances_open_problems}.
This upload cost can be a major bottleneck, especially when the model is large and the number of clients is substantial.
To reduce communication costs, researchers have proposed various compression techniques, such as quantization~\cite{qsgd}, low-rank factorization~\cite{vogels2019powersgd}, sparsification~\cite{aji2017sparse}, and sketching~\cite{communication_efficient}, among others.
The download cost is generally not considered a bottleneck, since clients tend to have less upload than download bandwidth, and because the effects of averaging across many clients can enable more aggressive lossy compression schemes~\cite{advances_open_problems}.

However, achieving convergence guarantees with upload compression presents theoretical and practical challenges.
While theoretical analyses often rely on unbiased compression, practical systems favor biased methods due to their computational efficiency and superior performance~\cite{advances_open_problems,beznosikov2023biased}.
To match the theoretical convergence properties of unbiased approaches, and to converge in practice in aggressive regimes, biased compression needs error feedback (also known as error compensation)~\cite{beznosikov2023biased,error_feedback}.
This mechanism requires the server to maintain client-specific control variates that track the state of each client, which creates significant limitations across distributed learning scenarios.
In privacy-focused applications like FL, server-side client tracking contradicts fundamental privacy principles.
Additionally, many distributed systems lack the infrastructure to maintain per-client state, and in massive cross-device deployments, clients are typically stateless~\cite{advances_open_problems}, making error feedback infeasible.

Motivated by the above challenges, we propose a novel distributed learning framework that allows highly compressible client updates without requiring control variates, which we call \algnamelong. 
Our framework leverages the previous aggregated update at the server to help clients compute a more compressible local update.
Clients compress the difference between their local update and the previous aggregated update, and the server adds the previous aggregated update when decoding the received messages.
Note that clients must receive the previous aggregated update along with the updated model, thus potentially doubling the download cost.
However, server-to-client communication is often cheap, as in distributed learning settings, it is primarily the clients who are resource constrained~\cite{advances_open_problems}.
This approach is inspired by error feedback, but does not require control variates, making it compatible with existing privacy mechanisms in FL and suitable for stateless clients.
The idea of compressing the compensated errors, for example in motion compensation and temporal prediction, is widely used in video coding~\cite{993440}.


\section{Related Work}\label{sec:related_work}
Communication compression is a well-studied topic in distributed learning, and error feedback is often suggested to improve convergence guarantees~\cite{SCAFFOLD}.
In~\cite{error_feedback}, the authors study the error feedback mechanism for one-bit per coordinate biased compression.
For general sparse compressors, it was studied in~\cite{sparsifiedSGD,sparsifiedGradientMethods}.
For the decentralized setting,~\cite{chocosgd,ortegaGossip} proposed variants of error feedback with general compression operators.
For asynchronous methods,~\cite{QAFeLworkshop,ortega2024quantized} also showed that a modified error feedback with general compression operators has good convergence guarantees.
In the non-convex setting,~\cite{EF21} showed that error feedback can be used in arbitrarily heterogeneous settings, which was later extended to the stochastic and convex settings in~\cite{beznosikov2023biased}.

\section{\algname{} overview}\label{sec:method}

To discuss the algorithm design, first, we must cover some compression preliminaries.
When clients send a message to the server, they first encode it using a function $E$.
The server decodes the received information using a function $D$.
We call these functions the encoder and decoder, respectively.
For a general compression mechanism, the composition $D(E(x)) := \compress{x}$ is called a compression operator~\cite{sparsifiedSGD}.
\begin{definition}
    A compression operator is a function $\mathcal{C} : \R^d \to \R^d$, paired with a positive compression parameter $\omega < 1$, such that for any vector $x$,
    \begin{equation}\label{eq:compression}
        \ec{\sqn{\compress{x} - x}} \leq \omega \sqn{x}.
    \end{equation}
\end{definition}
\begin{example}[Top-k compression]\label{ex:topk}
    The top-k compression operator sets all but the top $k$ elements of a vector in absolute value to zero. The top-k compression operator has parameter $\omega = 1 - \frac{k}{d}$~\cite{qsgd}.
\end{example}

Next, we describe how compression operators are used when minimizing the global loss function from \Cref{eq:global_loss} in a distributed learning setting.
The fundamental algorithm for this purpose is Distributed Compressed Gradient Descent (DCGD) --- see \Cref{alg:classic-gd}.
The pseudocode shows how, at each round, the global model is sent to the clients, which train it using gradients computed with local data.
Clients then compress these gradients and send them to the server, which averages them to update the global model.
This process is repeated for any desired number of rounds.
\begin{algorithm}[htbp]
    \caption{Distributed Compressed Gradient Descent}\label{alg:classic-gd}
    \begin{algorithmic}[1]
        \State{} \textbf{Input:} Global model $x$,  Rounds $K$, Encoder-Decoder $(E,D)$ pair for compression, learning rate $\gamma$
        \State{} Initialize global model $x^0$, and aggregate $\Delta_s^0 \gets 0$
        \For{round $k$ from 1 to $K$}
        \State{} Send $x^k$ to all clients
        \For{each client $n$ in parallel}
        \State{} $y_n^k \gets x^k - \gamma \nabla f_n(x^k)$ \Comment{Train $x^k$ using local data, store the output in $y_n^k$}
        \State{} $\Delta_n^k \gets y_n^k - x^k =  - \gamma \nabla f_n(x^k)$ \Comment{Compute local update}
        \State{} Send $E(\Delta_n^k)$ to server \Comment{Upload local update}
        \EndFor{}
        \State{} Server decodes each client $n$ via $q_n^k \gets D(E(\Delta_n^k))$
        \State{} Aggregate client updates in $\Delta_s^k := \frac{1}{N} \sum q_n^k$
        \State{} Obtain $x^{k+1} := x^k + \Delta_s^k$.
        \EndFor{}
    \end{algorithmic}
\end{algorithm}
Note that DCGD is a specific instance of the general distributed learning framework, where we have chosen gradient descent as the optimizer for the local models, and equal-weight averaging for the aggregation strategy.
We can derive a general strategy by not determining the aggregation strategy for client updates, nor the optimizer for on-client training.

Our framework, \algname, leverages the previous aggregated update $\Delta_s^{k-1}$ to help clients compute a more compressible update.
Namely, clients will compress the difference between their local update $\Delta_n^k$ and the previous aggregated update:
\begin{align*}
    E(\Delta_n^k - \Delta_s^{k-1}).
\end{align*}
On the server side, the server will add the previous aggregated update when decoding the received messages:
\begin{align}
    q_n^k \gets D(E(\Delta_n^k - \Delta_s^{k-1})) + \Delta_s^{k-1}. \label{eq:cafe-decoding}
\end{align}
The pseudocode for this procedure is described in \Cref{alg:proposed}, where the novelty with respect to the general distributed learning framework is highlighted in green boxes.
\begin{algorithm}[htbp]
    \caption{\algname}\label{alg:proposed}
    \begin{algorithmic}[1]
        \State{} \textbf{Input:} Global model $x$, Rounds $K$, Encoder-Decoder $(E,D)$ pair for compression
        \State{} Initialize global model $x^0$, and aggregate $\Delta_s^0 \gets 0$
        \For{round $k$ from 1 to $K$}
        \State{} Send $x^k$ \mybox{and $\Delta_s^{k-1}$} to all clients \Comment{In the stateful version $\Delta_s^{k-1}$ may be omitted}
        \For{each client $n$ in parallel}
        \State{} $y_n^k \gets \text{Train}(x_n^k)$ \Comment{Train $x^k$ using local data, store the output in $y_n^k$}
        \State{} $\Delta_n^k \gets y_n^k - x^k$ \Comment{Compute local update}
        \State{} Send \mybox{$E(\Delta_n^k - \Delta_s^{k-1})$} to server \Comment{Upload difference}
        \EndFor{}
        \State{} Server decodes each client $n$ via \mybox{\cref{eq:cafe-decoding}}
        \State{} Aggregate client updates in $\Delta_s^k$
        \State{} Obtain $x^{k+1}$ using $x^k$ and $\Delta_s^k$
        \EndFor{}
    \end{algorithmic}
\end{algorithm}
Note that the error feedback mechanism in~\cite{EF21} is a special case of \algname{} with a single client. In this case, the aggregated update at the server is simply the client update, and we can analyze it as a control variate.
However, in the multi-client setting, the aggregated update is a combination of all client updates, which acts as a proxy for client-specific control variates and requires novel analysis, shown in \Cref{sec:analysis}.

Observe that if clients have memory, they can retain $x^{k-1}$.
In many popular distributed learning algorithms, $x^k$ and $x^{k-1}$ determine $\Delta_s^{k-1}$, like Distributed Gradient Descent, FedAvg, etc.
This means that the \algname{}'s server does not need to send $\Delta_s^{k-1}$ if clients have memory.
Algorithms with momentum can also easily be adapted to our framework.

\section{Analysis}\label{sec:analysis}
We analyze \algname{} using Gradient Descent as the optimizer of choice and a compression operator $\mathcal{C}$ with parameter $\omega < 1$.
We proceed with the following standard assumptions~\cite{advances_open_problems,SCAFFOLD}:
\begin{assumption}\label{as:L-smooth}
    The objective function $f$ is $L$-smooth, which implies that it is differentiable, $\nabla f$ is $L$-Lipschitz, and
    \begin{equation}
        f(y) \leq f(x) + \ev{\nabla f(x), y - x} + \frac{L}{2} \sqn{y - x}.
    \end{equation}
    Also, the objective function $f$ is lower-bounded by $f^\star$.
\end{assumption}
\begin{assumption}\label{as:bounded-dissimilarity}
    The local gradients have bounded dissimilarity, that is, there exists a $B^2 \geq 1$ such that
    \begin{align}
        \frac{1}{N} \sum \sqn{\nabla f_n (x)} \leq B^2 \sqn{\nabla f(x)}.
    \end{align}
\end{assumption}

We present the main results for DCGD without \algname{} (\Cref{thm:theorem-dcgd}), and with \algname{} (\Cref{thm:theorem-proposed}).
Please see~\Cref{appendix:proofs} for the proofs.
\begin{restatable}{theorem}{thmDcgd}\label{thm:theorem-dcgd}
    Given \Cref{as:bounded-dissimilarity,as:L-smooth}, a positive learning rate $\gamma$ such that $\gamma \leq \frac{1}{L}$, and a compression parameter $\omega < 1$, DCGD iterating over $K$ iterations satisfies
    \begin{align}
        \frac{1}{K} \sum_{k=0}^{K-1} \ec{\sqn{\nabla f(x^k)}} & \leq \frac{2 F_0}{\gamma K \br{1-\omega B^2}},
    \end{align}
    where $F_0 = f(x^0) - f^\star$, as long as $1 > \omega B^2$.
\end{restatable}
\begin{corollary}\label{cor:corollary-dcgd}
    Given \Cref{as:bounded-dissimilarity,as:L-smooth}, a compression parameter $\omega < 1$, and $\gamma = 1/L$, DCGD over $K$ iterations results in the following upper bound:
    \begin{align}
        \frac{1}{K} \sum_{k=0}^{K-1} \ec{\sqn{\nabla f(x^k)}} & \leq \frac{2 L F_0}{K \br{1-\omega B^2}},
    \end{align}
    where $F_0 = f(x^0) - f^\star$.
\end{corollary}

\begin{restatable}{theorem}{thmCAFe}\label{thm:theorem-proposed}
    Given \Cref{as:bounded-dissimilarity,as:L-smooth}, a positive learning rate $\gamma$ such that
    \begin{align}
        \gamma \leq \frac{1-\omega}{L \br{1+\omega}}\label{eq:gamma-condition},
    \end{align}
    \algname{} + DGD iterating over $K$ iterations results in
    \begin{align}
        \frac{1}{K} \sum_{k=0}^{K-1} \ec{\sqn{\nabla f(x^k)}} & \leq \frac{2 F_0 \br{1-\omega}}{\gamma K \br{1-\omega B^2}},
    \end{align}
    where $F_0 = f(x^0) - f^\star$, as long as $1 > \omega B^2$.
\end{restatable}
\begin{corollary}\label{cor:corollary-proposed}
    Given \Cref{as:bounded-dissimilarity,as:L-smooth}, a compression parameter $\omega < 1$, and $\gamma = \frac{1-\omega}{L \br{1+\omega}}$, \algname{} + DGD over $K$ iterations results in the following upper bound:
    \begin{align}
        \frac{1}{K} \sum_{k=0}^{K-1} \ec{\sqn{\nabla f(x^k)}} & \leq \frac{2 L F_0 \br{1+\omega} }{K \br{1-\omega B^2}},
    \end{align}
    where $F_0 = f(x^0) - f^\star$.
\end{corollary}
Observing \Cref{thm:theorem-dcgd,thm:theorem-proposed}, given a choice of learning rate that satisfies both assumptions, \algname{} + DGD improves the convergence rate of DCGD by a factor of $(1-\omega)$.
This can be a significant improvement when the compression parameter is close to 1, which is the case for aggressive compression.

If the learning rate is tuned separately for each approach to be the largest possible, the DCGD's upper bound is smaller than \algname{} + DGD's, as per \Cref{cor:corollary-dcgd,cor:corollary-proposed}.
However, this is a very aggressive choice of learning rate, and in practice, it is unlikely to be chosen.
Also, the difference is a factor $\br{1+\omega} < 2$, which is negligible in most cases.

\section{Experimental Results}\label{sec:experiments}
\begin{figure}[htbp]
    \vspace{-.3cm}
    \centering
    \subfloat[MNIST --- \emph{iid}]{
    \hspace{-.3cm}\includegraphics[width=0.255\textwidth]{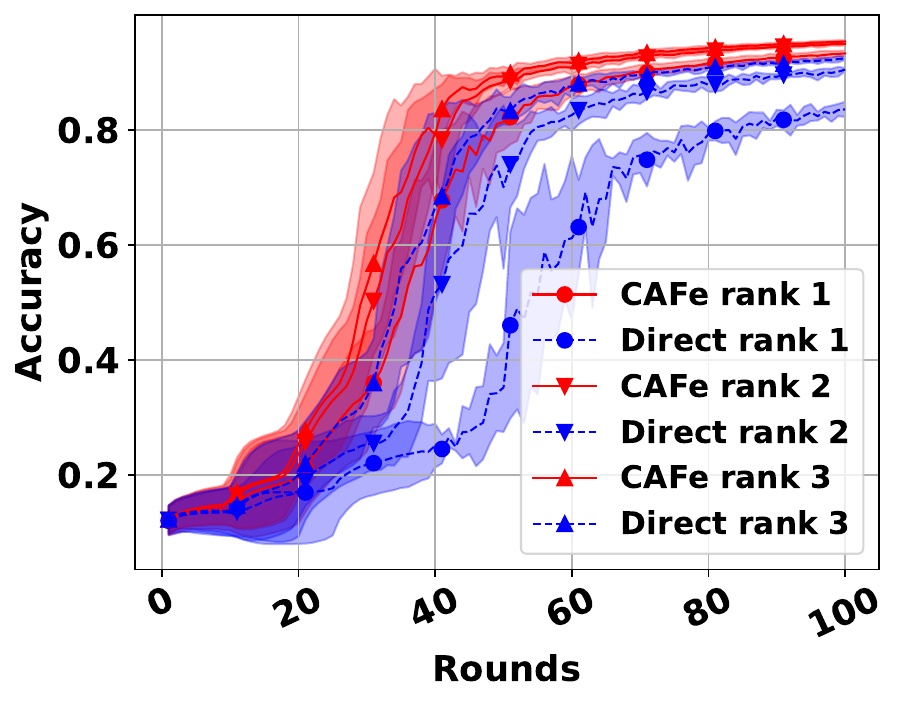}
    }
    \subfloat[MNIST --- \emph{non-iid}]{
    \hspace{-.3cm}\includegraphics[width=0.255\textwidth]{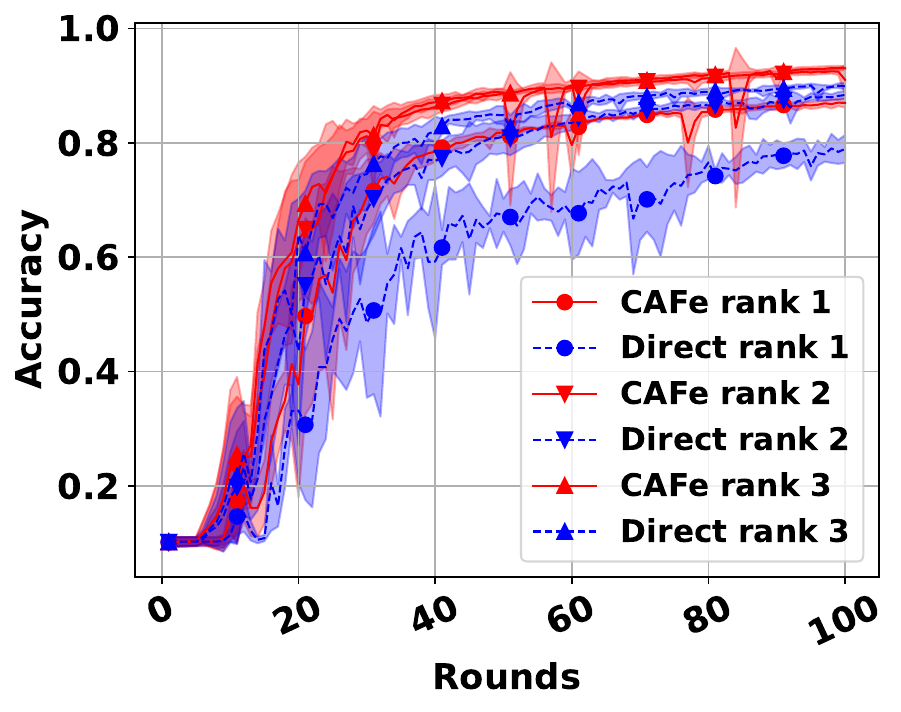}
    }
    \\
    \vspace{-.3cm}
    \subfloat[EMNIST --- \emph{iid}]{
    \hspace{-.3cm}\includegraphics[width=0.255\textwidth]{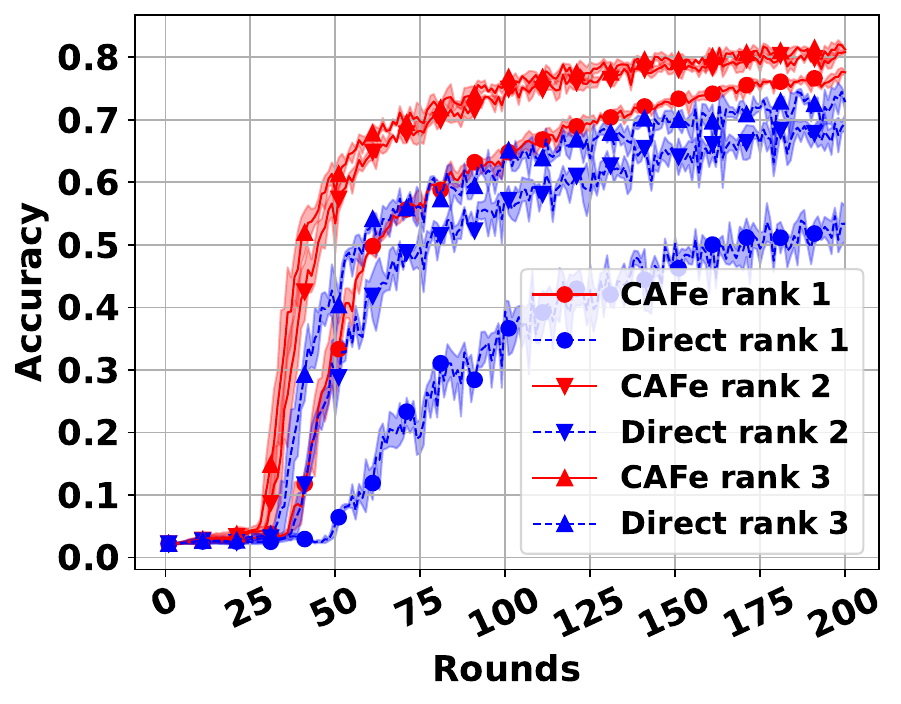}
    }
    \subfloat[EMNIST --- \emph{non-iid}]{
    \hspace{-.3cm}\includegraphics[width=0.255\textwidth]{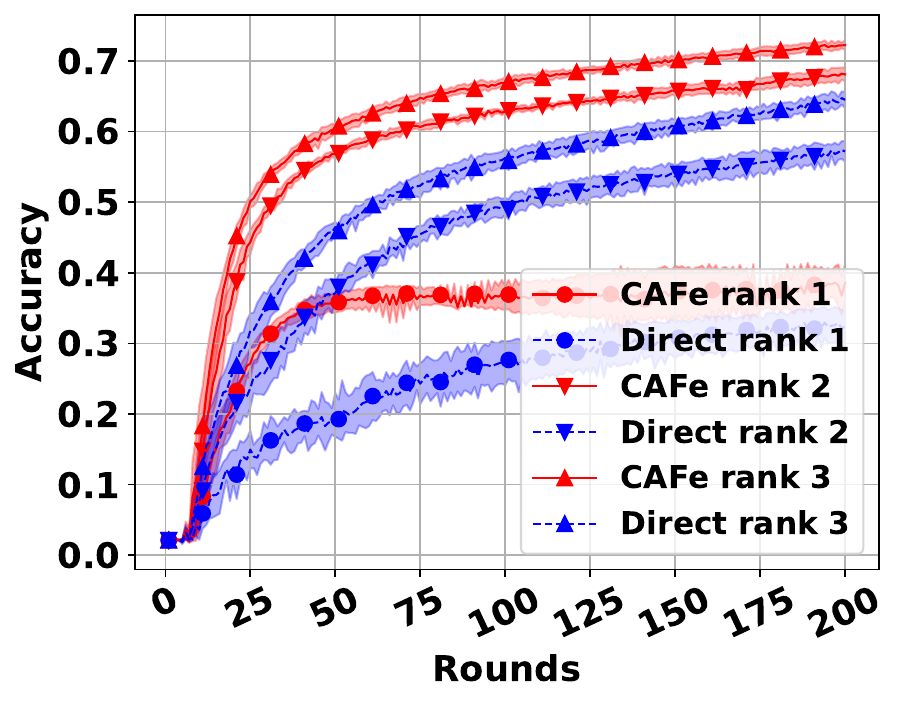}
    }
    \\
    \vspace{-.3cm}
    \subfloat[CIFAR100 --- \emph{iid}]{
    \hspace{-.3cm}\includegraphics[width=0.255\textwidth]{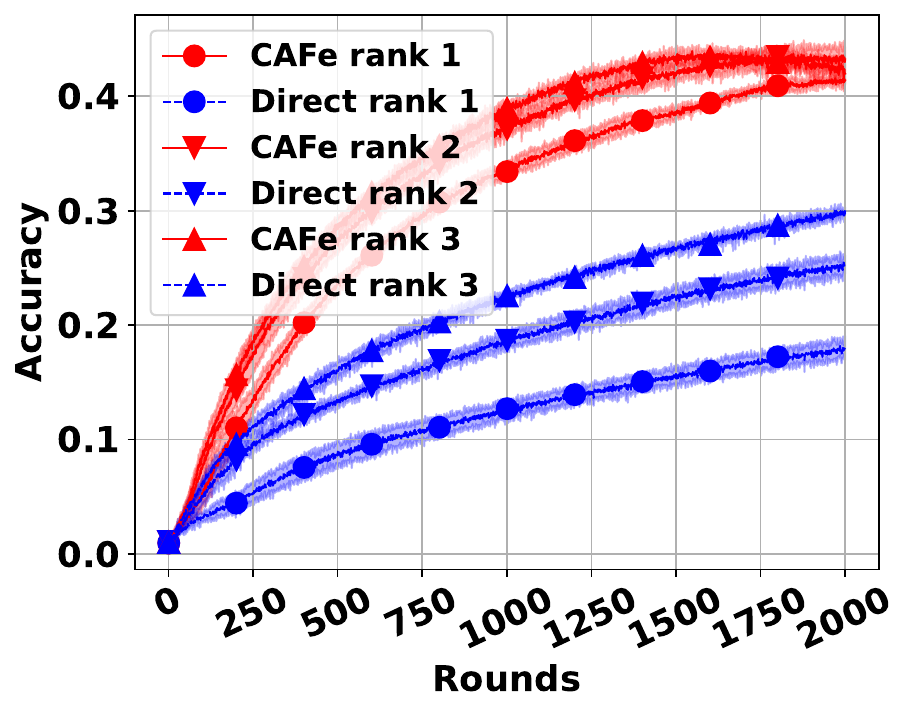}
    }
    \subfloat[CIFAR100 --- \emph{non-iid}]{
    \hspace{-.3cm}\includegraphics[width=0.255\textwidth]{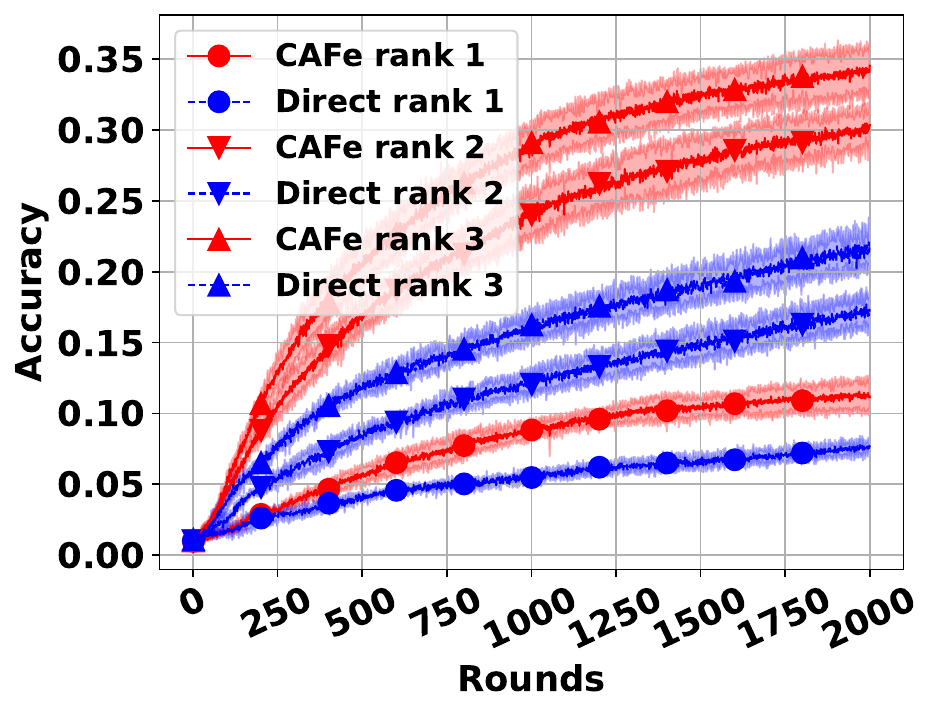}
    }
    \vspace{-.1cm}
    \caption{SVD compression performance on MNIST, EMNIST, and CIFAR-100. (a, c, e): \emph{iid} setting. (b, d, f): \emph{non-iid} setting.
    }\label{fig:perf}
\vspace{-.1cm}
\end{figure}

\begin{table*}[htbp]
\centering
\caption{Comparison between \algname{} and Direct compression under 4 compression methods with various parameter settings.}
\vspace{-.15cm}
\resizebox{\textwidth}{!}{
\begin{tabular}{|c|c|c|c|c|c|c|c|c|c|c|}
\hline
\multicolumn{11}{|c|}{\textbf{Top-k (top $10\%$, $1\%$, or $0.1\%$ of coordinates)}}\\ 
\hline
\multicolumn{2}{|c|}{    }                                                  & \multicolumn{3}{|c|}{\textbf{MNIST}}                                                         & \multicolumn{3}{|c|}{\textbf{EMNIST}}                                                        & \multicolumn{3}{|c|}{\textbf{CIFAR-100}}                                 \\
\hline
\multicolumn{2}{|c|}{\textbf{k}}                                                       & \multicolumn{1}{|c|}{0.1} & \multicolumn{1}{|c|}{0.01} & \multicolumn{1}{|c|}{0.001} & \multicolumn{1}{|c|}{0.1} & \multicolumn{1}{|c|}{0.01} & \multicolumn{1}{|c|}{0.001} & \multicolumn{1}{|c|}{0.1} & \multicolumn{1}{|c|}{0.01} & 0.001 \\ \hline
\multicolumn{1}{|c|}{\multirow{2}{*}{\textit{iid}}}     & \multicolumn{1}{|c|}{\textbf{Direct}} & \multicolumn{1}{|c|}{$95.51 \pm 0.30$}    & \multicolumn{1}{|c|}{$92.99 \pm 0.53$}     & \multicolumn{1}{|c|}{$44.99 \pm 3.12$}      & \multicolumn{1}{|c|}{$81.27 \pm 7.49$}    & \multicolumn{1}{|c|}{$77.95 \pm 3.11$}     & \multicolumn{1}{|c|}{$71.63 \pm 1.34$}      & \multicolumn{1}{|c|}{$\mathbf{39.44 \pm 0.66}$}    & \multicolumn{1}{|c|}{$30.60 \pm 1.86$}     &   $16.48 \pm 0.35$    \\ \cline{2-11} 
\multicolumn{1}{|c|}{}                         & \multicolumn{1}{|c|}{\textbf{\algname{}}}    & \multicolumn{1}{|c|}{$\mathbf{95.90 \pm 0.24}$}    & \multicolumn{1}{|c|}{$\mathbf{95.05 \pm 0.34}$}     & \multicolumn{1}{|c|}{$\mathbf{91.42 \pm 1.25}$}      & \multicolumn{1}{|c|}{$\mathbf{83.54 \pm 4.38}$}    & \multicolumn{1}{|c|}{$\mathbf{80.19 \pm 1.76}$}     & \multicolumn{1}{|c|}{$\mathbf{75.32 \pm 1.93}$}      & \multicolumn{1}{|c|}{$37.82 \pm 1.63$}    & \multicolumn{1}{|c|}{$\mathbf{37.10 \pm 1.96}$}     &   $\mathbf{20.54 \pm 1.73}$    \\ \hline
\multicolumn{1}{|c|}{\multirow{2}{*}{\textit{non-iid}}} & \multicolumn{1}{|c|}{\textbf{Direct}} & \multicolumn{1}{|c|}{$92.18 \pm 0.05$}    & \multicolumn{1}{|c|}{$89.96 \pm 0.37$}     & \multicolumn{1}{|c|}{$82.63 \pm 2.86$}      & \multicolumn{1}{|c|}{$73.54 \pm 0.15$}    & \multicolumn{1}{|c|}{$71.06 \pm 0.61$}     & \multicolumn{1}{|c|}{$60.33 \pm 0.82$}      & \multicolumn{1}{|c|}{$35.99 \pm 3.08$}    & \multicolumn{1}{|c|}{$\mathbf{24.34 \pm 1.49}$}     &    $\mathbf{10.17 \pm 0.91}$   \\ \cline{2-11} 
\multicolumn{1}{|c|}{}                         & \multicolumn{1}{|c|}{\textbf{\algname{}}}    & \multicolumn{1}{|c|}{$\mathbf{92.73 \pm 0.37}$}    & \multicolumn{1}{|c|}{$\mathbf{91.15 \pm 0.66}$}     & \multicolumn{1}{|c|}{$\mathbf{88.31 \pm 1.53}$}      & \multicolumn{1}{|c|}{$\mathbf{74.56 \pm 0.10}$}    & \multicolumn{1}{|c|}{$\mathbf{72.07 \pm 0.91}$}     & \multicolumn{1}{|c|}{$\mathbf{63.65 \pm 0.61}$}      & \multicolumn{1}{|c|}{$\mathbf{38.26 \pm 2.46}$}    & \multicolumn{1}{|c|}{$23.11 \pm 0.53$}     &    $7.27 \pm 1.47$   \\ 
\hline
\hline
\multicolumn{11}{|c|}{\textbf{Top-k (top $10\%$) + Quantization ($4,5,$ or $6$ bits per coordinate) }}\\ 
\hline
\multicolumn{2}{|c|}{    }                                                  & \multicolumn{3}{|c|}{\textbf{MNIST}}                                                         & \multicolumn{3}{|c|}{\textbf{EMNIST}}                                                        & \multicolumn{3}{|c|}{\textbf{CIFAR-100}}                                 \\
\hline
\multicolumn{2}{|c|}{\textbf{bits}}                                                       & \multicolumn{1}{|c|}{4} & \multicolumn{1}{|c|}{5} & \multicolumn{1}{|c|}{6} & \multicolumn{1}{|c|}{4} & \multicolumn{1}{|c|}{5} & \multicolumn{1}{|c|}{6} & \multicolumn{1}{|c|}{4} & \multicolumn{1}{|c|}{5} & 6 \\ \hline
\multicolumn{1}{|c|}{\multirow{2}{*}{\textit{iid}}}     & \multicolumn{1}{|c|}{\textbf{Direct}} & \multicolumn{1}{|c|}{$20.94 \pm 13.57$}    & \multicolumn{1}{|c|}{$92.54 \pm 1.49$}     & \multicolumn{1}{|c|}{$95.13 \pm 0.32$}      & \multicolumn{1}{|c|}{$18.65 \pm 22.92$}    & \multicolumn{1}{|c|}{$77.45 \pm 0.26$}     & \multicolumn{1}{|c|}{$80.90 \pm 0.97$}      & \multicolumn{1}{|c|}{$\mathbf{15.79 \pm 6.96}$}    & \multicolumn{1}{|c|}{$32.72 \pm 2.81$}     &    $\mathbf{38.04 \pm 0.66}$   \\ \cline{2-11} 
\multicolumn{1}{|c|}{}                         & \multicolumn{1}{|c|}{\textbf{\algname{}}}    & \multicolumn{1}{|c|}{$\mathbf{64.49 \pm 37.58}$}    & \multicolumn{1}{|c|}{$\mathbf{94.04 \pm 1.48}$}     & \multicolumn{1}{|c|}{$\mathbf{95.50 \pm 0.37}$}      & \multicolumn{1}{|c|}{$\mathbf{24.41 \pm 30.13}$}    & \multicolumn{1}{|c|}{$\mathbf{81.12 \pm 0.83}$}     & \multicolumn{1}{|c|}{$\mathbf{82.99 \pm 0.65}$}      & \multicolumn{1}{|c|}{$12.12 \pm 4.62$}    & \multicolumn{1}{|c|}{$\mathbf{33.80 \pm 5.94}$}     &    $33.59 \pm 0.93$   \\ \hline
\multicolumn{1}{|c|}{\multirow{2}{*}{\textit{non-iid}}} & \multicolumn{1}{|c|}{\textbf{Direct}} & \multicolumn{1}{|c|}{$\mathbf{11.35 \pm 0.00}$}    & \multicolumn{1}{|c|}{$\mathbf{36.36 \pm 35.37}$}     & \multicolumn{1}{|c|}{$\mathbf{88.79 \pm 2.03}$}      & \multicolumn{1}{|c|}{$63.82 \pm 4.34$}    & \multicolumn{1}{|c|}{$68.38 \pm 2.33$}     & \multicolumn{1}{|c|}{$72.42 \pm 0.65$}      & \multicolumn{1}{|c|}{$\mathbf{17.23 \pm 2.79}$}    & \multicolumn{1}{|c|}{$\mathbf{29.16 \pm 1.36}$}     &   $\mathbf{36.76 \pm 0.81}$    \\ \cline{2-11} 
\multicolumn{1}{|c|}{}                         & \multicolumn{1}{|c|}{\textbf{\algname{}}}    & \multicolumn{1}{|c|}{$\mathbf{11.35 \pm 0.00}$}    & \multicolumn{1}{|c|}{$35.94 \pm 38.78$}     & \multicolumn{1}{|c|}{$87.04 \pm 5.02$}      & \multicolumn{1}{|c|}{$\mathbf{70.06 \pm 1.97}$}    & \multicolumn{1}{|c|}{$\mathbf{72.10 \pm 1.02}$}     & \multicolumn{1}{|c|}{$\mathbf{73.89 \pm 0.81}$}      & \multicolumn{1}{|c|}{$5.23 \pm 3.05$}    & \multicolumn{1}{|c|}{$25.86 \pm 0.77$}     &    $34.78 \pm 2.47$   \\ 
\hline
\hline
\multicolumn{11}{|c|}{\textbf{SVD}} \\
\hline
\multicolumn{2}{|c|}{}                                                  & \multicolumn{3}{|c|}{\textbf{MNIST}}                                                         & \multicolumn{3}{|c|}{\textbf{EMNIST}}                                                        & \multicolumn{3}{|c|}{\textbf{CIFAR-100}}                                 \\
\hline
\multicolumn{2}{|c|}{\textbf{rank}}                                                       & \multicolumn{1}{|c|}{1} & \multicolumn{1}{|c|}{2} & \multicolumn{1}{|c|}{3} & \multicolumn{1}{|c|}{1} & \multicolumn{1}{|c|}{2} & \multicolumn{1}{|c|}{3} & \multicolumn{1}{|c|}{1} & \multicolumn{1}{|c|}{2} &  3 \\ \hline
\multicolumn{1}{|c|}{\multirow{2}{*}{\textit{iid}}}     & \multicolumn{1}{|c|}{\textbf{Direct}} & \multicolumn{1}{|c|}{$83.54 \pm 1.31$}    & \multicolumn{1}{|c|}{$90.45 \pm 2.90$}     & \multicolumn{1}{|c|}{$92.42 \pm 2.82$}      & \multicolumn{1}{|c|}{$53.38 \pm 2.83$}    & \multicolumn{1}{|c|}{$69.48 \pm 1.27$}     & \multicolumn{1}{|c|}{$72.84 \pm 1.28$}      & \multicolumn{1}{|c|}{$18.00 \pm 1.07$}    & \multicolumn{1}{|c|}{$25.30 \pm 0.70$}     &    $30.03 \pm 2.74$   \\ \cline{2-11} 
\multicolumn{1}{|c|}{}                         & \multicolumn{1}{|c|}{\textbf{\algname{}}}    & \multicolumn{1}{|c|}{$\mathbf{93.27 \pm 0.48}$}    & \multicolumn{1}{|c|}{$\mathbf{94.88 \pm 0.21}$}     & \multicolumn{1}{|c|}{$\mathbf{95.34 \pm 0.23}$}      & \multicolumn{1}{|c|}{$\mathbf{77.59 \pm 0.28}$}    & \multicolumn{1}{|c|}{$\mathbf{80.81 \pm 0.18}$}     & \multicolumn{1}{|c|}{$\mathbf{81.26 \pm 0.49}$}      & \multicolumn{1}{|c|}{$\mathbf{41.39 \pm 0.30}$}    & \multicolumn{1}{|c|}{$\mathbf{43.21 \pm 1.11}$}     &  $\mathbf{42.00 \pm 0.70}$     \\ \hline
\multicolumn{1}{|c|}{\multirow{2}{*}{\textit{non-iid}}} & \multicolumn{1}{|c|}{\textbf{Direct}} & \multicolumn{1}{|c|}{$78.93 \pm 2.44$}    & \multicolumn{1}{|c|}{$88.36 \pm 0.71$}     & \multicolumn{1}{|c|}{$89.92 \pm 0.57$}      & \multicolumn{1}{|c|}{$32.44 \pm 2.08$}    & \multicolumn{1}{|c|}{$57.18 \pm 1.45$}     & \multicolumn{1}{|c|}{$64.47 \pm 1.13$}      & \multicolumn{1}{|c|}{$7.63 \pm 0.68$}    & \multicolumn{1}{|c|}{$17.19 \pm 1.22$}     &   $21.72 \pm 1.36$    \\ \cline{2-11} 
\multicolumn{1}{|c|}{}                         & \multicolumn{1}{|c|}{\textbf{\algname{}}}    & \multicolumn{1}{|c|}{$\mathbf{87.00 \pm 0.83}$}    & \multicolumn{1}{|c|}{$\mathbf{91.02 \pm 2.58}$}     & \multicolumn{1}{|c|}{$\mathbf{93.04 \pm 0.35}$}      & \multicolumn{1}{|c|}{$\mathbf{38.57 \pm 2.65}$}    & \multicolumn{1}{|c|}{$\mathbf{68.10 \pm 0.98}$}     & \multicolumn{1}{|c|}{$\mathbf{72.23 \pm 0.65}$}      & \multicolumn{1}{|c|}{$\mathbf{11.14 \pm 1.03}$}    & \multicolumn{1}{|c|}{$\mathbf{30.35 \pm 1.70}$}     &   $\mathbf{34.54 \pm 1.55}$    \\ 
\hline
\hline
\multicolumn{11}{|c|}{\textbf{SVD (rank $1$) + Quantization ($4,5,$ or $6$ bits per coordinate)}}\\ 
\hline
\multicolumn{2}{|c|}{}                                                  & \multicolumn{3}{|c|}{\textbf{MNIST}}                                                         & \multicolumn{3}{|c|}{\textbf{EMNIST}}                                                        & \multicolumn{3}{|c|}{\textbf{CIFAR-100}}                                 \\
\hline
\multicolumn{2}{|c|}{\textbf{bits}}                                                       & \multicolumn{1}{|c|}{2} & \multicolumn{1}{|c|}{3} & \multicolumn{1}{|c|}{4} & \multicolumn{1}{|c|}{2} & \multicolumn{1}{|c|}{3} & \multicolumn{1}{|c|}{4} & \multicolumn{1}{|c|}{2} & \multicolumn{1}{|c|}{3} &  4 \\ \hline
\multicolumn{1}{|c|}{\multirow{2}{*}{\textit{iid}}}     & \multicolumn{1}{|c|}{\textbf{Direct}} & \multicolumn{1}{|c|}{$68.94 \pm 26.21$}    & \multicolumn{1}{|c|}{$85.84 \pm 0.99$}     & \multicolumn{1}{|c|}{$83.59 \pm 0.71$}      & \multicolumn{1}{|c|}{$35.18 \pm 16.71$}    & \multicolumn{1}{|c|}{$52.77 \pm 3.04$}     & \multicolumn{1}{|c|}{$51.15 \pm 1.54$}      & \multicolumn{1}{|c|}{$12.78 \pm 0.49$}    & \multicolumn{1}{|c|}{$16.81 \pm 1.39$}     &    $18.43 \pm 0.17$   \\ \cline{2-11} 
\multicolumn{1}{|c|}{}                         & \multicolumn{1}{|c|}{\textbf{\algname{}}}    & \multicolumn{1}{|c|}{$\mathbf{90.45 \pm 0.46}$}    & \multicolumn{1}{|c|}{$\mathbf{92.77 \pm 0.41}$}     & \multicolumn{1}{|c|}{$\mathbf{93.29 \pm 0.42}$}      & \multicolumn{1}{|c|}{$\mathbf{55.67 \pm 5.97}$}    & \multicolumn{1}{|c|}{$\mathbf{72.43 \pm 1.11}$}     & \multicolumn{1}{|c|}{$\mathbf{77.19 \pm 0.36}$}      & \multicolumn{1}{|c|}{$\mathbf{20.28 \pm 2.91}$}    & \multicolumn{1}{|c|}{$\mathbf{32.57 \pm 2.74}$}     &  $\mathbf{38.55 \pm 0.92}$     \\ \hline
\multicolumn{1}{|c|}{\multirow{2}{*}{\textit{non-iid}}} & \multicolumn{1}{|c|}{\textbf{Direct}} & \multicolumn{1}{|c|}{$\mathbf{66.51 \pm 2.65}$}    & \multicolumn{1}{|c|}{$79.73 \pm 0.75$}     & \multicolumn{1}{|c|}{$79.22 \pm 2.09$}      & \multicolumn{1}{|c|}{$\mathbf{14.31 \pm 4.15}$}    & \multicolumn{1}{|c|}{$30.11 \pm 2.71$}     & \multicolumn{1}{|c|}{$3.73 \pm 2.54$}      & \multicolumn{1}{|c|}{$4.66 \pm 0.04$}    & \multicolumn{1}{|c|}{$7.53 \pm 0.48$}     &   $7.78 \pm 0.94$    \\ \cline{2-11} 
\multicolumn{1}{|c|}{}                         & \multicolumn{1}{|c|}{\textbf{\algname{}}}    & \multicolumn{1}{|c|}{$63.33 \pm 10.19$}    & \multicolumn{1}{|c|}{$\mathbf{86.33 \pm 0.30}$}     & \multicolumn{1}{|c|}{$\mathbf{86.96 \pm 0.37}$}      & \multicolumn{1}{|c|}{$6.23 \pm 5.80$}    & \multicolumn{1}{|c|}{$\mathbf{41.76 \pm 0.65}$}     & \multicolumn{1}{|c|}{$\mathbf{40.12 \pm 1.51}$}      & \multicolumn{1}{|c|}{$\mathbf{11.41 \pm 0.42}$}    & \multicolumn{1}{|c|}{$\mathbf{12.52 \pm 1.08}$}     &   $\mathbf{12.01 \pm 1.12}$    \\ 
\hline
\end{tabular}
}\label{tab:exp_result}
\vspace{-.3cm}
\end{table*}

We present FL experiments with 10 clients.
The selected datasets are MNIST
, EMNIST
, and CIFAR-100
, and we follow~\cite{isik2024adaptive} to choose models for the three datasets, which are \texttt{CONV4}, \texttt{CONV4}, and ResNet-18, respectively.
The learning rates are tuned based on the model architectures to be as large as possible without model divergence.
Experimentally, we find them to be the same for DCGD and \algname{}.
We present results for both homogeneous and heterogeneous data cases, denoted \textit{iid} and \textit{non-iid}, respectively.
For the latter, we randomly sample $40\%$ of total classes for each client.
We perform one local training epoch with batch size 512 and vary the number of global training rounds for each experiment.
Please see~\Cref{appendix:experimental_setup} for the experimental setup.
We run each experiment with 3 random seeds and report the final accuracy means and standard deviations.
We show the effectiveness of \algname{} compared with direct compression using the following four biased compression methods: Top-k (see~\Cref{ex:topk}), Top-k + Quantization, Singular Value Decomposition (SVD)~\cite{vogels2019powersgd}, and SVD + Quantization under various compression parameter settings, as reported in Table~\ref{tab:exp_result}.
Sparsification is performed before quantization since it is optimal for FL~\cite{harma2024effective}.
The results align with our theory:
\algname{} outperforms existing direct compression methods in moderate heterogeneity settings (MNIST and EMNIST, iid settings), while it may suffer when the heterogeneity is higher and compression is very aggressive (CIFAR-100, select non-iid settings).
Since SVD provides a high level of compression with a low bitrate, we show the convergence rates by plotting the learning curves using SVD compression in Fig.~\ref{fig:perf}.
Observe that not only does \algname{} achieve better performance, but also, compared to direct compression, it converges faster.
\vspace{-.3cm}
\section{Conclusion}\label{sec:conclusion}
We proposed \algnamelong, a novel framework for bandwidth-efficient distributed learning.
By leveraging the previous aggregated update, \algname{} makes local updates more compressible, reducing upload costs for biased compressors. We proved convergence guarantees when optimizing locally with Gradient Descent and demonstrated experimentally that \algname{} outperforms direct compression for compressors used in practice.

\bibliographystyle{IEEEbib}
\bibliography{references}

\appendix
\section{Additional Experiment Details}\label{appendix:experimental_setup}
As reported in Table~\ref{tab:exp_result}, we select $k=10\%, 1\%, \mbox{and } 0.1\%$ for Top-k methods.
We also choose uniform quantization with 4, 5, and 6 bits for Top-k + Quantization and uniform quantization with 2, 3, and 4 bits for SVD + Quantization.
For our Quantization experiments, we fix k to $10\%$ for Top-k and rank to $1$ for SVD.
For Top-k + Quantization compression, we aim to test the lower limit of the choice for the number of bits per coordinate.
Notice in~\Cref{tab:exp_result} that when the number of bits is less than 5, both \algname{} and direct compression result in low performance and large variance, which indicates that $n_{bits}>5$ is more suitable for this compression.
For SVD compression, \algname{} consistently outperforms direct compression by a large margin, regardless of the model architecture and dataset.
This is due to SVD's low compression error, even when using rank 1.
With SVD + Quantization, we also test the lower limit and find that it is suitable to choose $n_{bits}>2$.
\begin{table}[htbp]
    \centering
    \caption{Experiment setup.}
    \resizebox{\columnwidth}{!}{%
        \begin{tabular}{|c|c|c|c|}
            \hline
                                                                         & \textbf{MNIST} & \textbf{EMNIST} & \textbf{CIFAR-100} \\ \hline
            \multicolumn{1}{|c|}{\textbf{Model}}                         & \texttt{CONV4} & \texttt{CONV4}  & ResNet-18          \\ \hline
            \multicolumn{1}{|c|}{\textbf{Learning Rate}}                 & 0.01           & 0.01            & 0.1                \\ \hline
            \multicolumn{1}{|c|}{\textbf{\# classes (\textit{non-iid})}} & 4              & 4               & 40                 \\ \hline
            \multicolumn{1}{|c|}{\textbf{FL Rounds}}                     & 100            & 200             & 2000               \\ \hline
        \end{tabular}
    }\label{tab:exp_setup}
\end{table}

\section{Proofs}\label{appendix:proofs}
We analyze \algname{} using Gradient Descent as the optimizer of choice and a compression operator $\mathcal{C}$ with parameter $\omega < 1$.
In this case, the iterates of \algname{} are:
\begin{align}
    x^{k+1}    & = x^{k} + \Delta_s^{k},                                                     \\
    \Delta_s^k & = \frac{1}{N}\sum_n \compress{\Delta_n^k - \Delta_s^{k-1}} + \Delta_s^{k-1} \\
    \Delta_n^k & = - \gamma \nabla f_n(x^k).
\end{align}
We define $e_n^k = \compress{\Delta_n^k - \Delta_s^{k-1}} - \br{\Delta_n^k - \Delta_s^{k-1}}$ as the compression error, and $\hat e_n^k = \frac{e_n^k}{\gamma}$ as the re-scaled compression error.
Then, we obtain
\begin{align}
    \Delta_s^k & = \frac{1}{N}\sum_n \br{\Delta_n^k  + e_n^k} \\
    \Delta_n^k & = - \gamma (\nabla f_n (x^k) + \hat e^k_n).
\end{align}
Furthermore, we define the average re-scaled compression error as $\overline{e}^k := \frac{1}{N} \sum \hat e_n^k$.
Combining these equations, we obtain
\begin{align}\label{eq:iteration}
    x^{k+1} = x^{k} - \gamma \br{\nabla f (x^k) + \overline{e}^k}.
\end{align}
Observe that if we have a perfect compressor $\mathcal{C}$, that is, the compression error is zero, we recover Distributed Gradient Descent.
For ease of notation, we will denote $g^k = \nabla f(x^k) + \overline{e^k}$.
Therefore, the iterates of \algname{} are $x^{k+1} = x^k - \gamma g^k$.

Given \Cref{as:L-smooth}, and using the fact that $-\ev{a,b} = \frac{-\sqn{a} -\sqn{b} + \sqn{a - b}}{2}$, we can ensure

\begin{align}
    f(x^{k+1}) & \leq f(x^k) - \frac{\gamma}{2} \sqn{\nabla f(x^k)} + \frac{\gamma}{2} \sqn{\nabla f(x^k) - g^k}\nonumber \\
               & - \br{\frac{1}{2\gamma} - \frac{L}{2}} \sqn{x^{k+1} - x^k} .\label{eq:main-recursion}
\end{align}

The second term on the RHS represents the compression error $\sqn{\overline{e}^k}$, and we will bound it differently for DCGD and \algname.
We present the main results for DCGD without \algname{} (\Cref{thm:theorem-dcgd}), and with \algname{} (\Cref{thm:theorem-proposed}), restated below for clarity.
\thmDcgd* 
\begin{proof}
    The compression error for \Cref{alg:classic-gd} satisfies
    \begin{align}
        \ec{\sqn{\overline{e}^k}} & = \frac{1}{N}\sum_n \ec{\sqn{\hat e_n^k}}           \\
                                  & \leq \omega \frac{1}{N}\sum_n \sqn{\nabla f_n(x^k)} \\
                                  & \leq \omega B^2 \sqn{\nabla f(x^k)},
    \end{align}
    where we have used Jensen's inequality in the first step, \Cref{eq:compression} for the compression parameter, and \Cref{as:bounded-dissimilarity} in the last step.

    If we assume that $\gamma \leq \frac{1}{L}$, we can simplify \Cref{eq:main-recursion} to
    \begin{align*}
        \ec{f(x^{k+1})} \leq f(x^0) - \frac{\gamma}{2} \br{1-\omega B^2} \sum_{\ell = 0}^k \ec{\sqn{\nabla f(x^\ell)}},
    \end{align*}
    where we have telescoped the recursion for $k$ iterations.

    Averaging over $K$ iterations and re-arranging, we obtain the theorem's statement.
\end{proof}
To analyze DGD with \algname{}, we need the following preliminary lemmas.
\begin{lemma}\label{lem:inner-product}
    Given an $L$-smooth function $f$, and iterations of the form $x^{k+1} = x^k - \gamma g^k$, we have
    \begin{align}
        -\ev{\nabla f(x^{k+1}), g^k} \leq -\ev{\nabla f(x^{k}), g^k} + \gamma L \sqn{g^k}.
    \end{align}
\end{lemma}
\begin{proof}
    We have
    $\ev{\nabla f(x^k), g^k} -\ev{\nabla f(x^{k+1}), g^k} = \\
        \ev{\nabla f(x^{k+1}) - \nabla f(x^k), g^k}$,
    and this can be bounded by
    \begin{align*}
        \ev{\nabla f(x^{k+1}) - \nabla f(x^k), g^k} & \leq \norm{\nabla f(x^{k+1}) - \nabla f(x^k)} \norm{g^k} \\
                                                    & \leq \gamma L \sqn{g^k},
    \end{align*}
    where we have used the $L$-smoothness of $f$ in the last step. Re-arranging, we obtain the desired result.
\end{proof}

\begin{lemma}\label{lem:compression}
    Given \Cref{as:bounded-dissimilarity,as:L-smooth}, the compression error for DGD + \algname{} satisfies
    \begin{equation}
        \begin{aligned}
            \ec{\sqn{\overline{e}^{k+1}}} & \leq \omega  \ec{B^2 \sqn{\nabla f(x^{k+1})} - \sqn{\nabla f(x^k)}} \\
                                          & + \gamma 2\omega L \ec{\sqn{g^k}} + \omega \sqn{\overline{e}^k}.
        \end{aligned}
    \end{equation}
\end{lemma}
\begin{proof}
    \begin{align*}
        \ec{\sqn{\overline{e}^{k+1}}} & \leq \frac{1}{N} \sum_n\ec{\sqn{\hat e_n^{k+1}}}                                        \\
                                      & \leq \frac{\omega}{N} \sum_n \ec{ \sqn{\nabla f_n (x^{k+1}) - g^k}}                     \\
                                      & = \frac{\omega}{N} \sum_n \ec{ \sqn{\nabla f_n (x^{k+1}) \pm \nabla f(x^{k+1}) - g^k}}.
    \end{align*}
    We can bound the obtained sum by
    \begin{align*}
        \omega\ec{\br{B^2 - 1} \sqn{\nabla f(x^{k+1})} + \sqn{\nabla f(x^{k+1}) - g^k}},
    \end{align*}
    since the interior product term is null and we can bound the sum of square client gradients using \Cref{as:bounded-dissimilarity}.
    Now, the last term can be bounded using \Cref{lem:inner-product}, since
    \begin{align*}
        \sqn{\nabla f(x^{k+1}) - g^k} & = \sqn{\nabla f(x^{k+1})} + \sqn{g^k}                    \\
                                      & \quad - 2\ev{\nabla f(x^{k+1}), g^k}                     \\
                                      & \leq \sqn{\nabla f(x^{k+1})} + \sqn{g^k}                 \\
                                      & \quad -2\ev{\nabla f (x^k), g^k} + 2\gamma L \sqn{g^k}   \\
                                      & = \sqn{\nabla f(x^{k+1})} - \sqn{\nabla f(x^k)}          \\
                                      & \quad + \sqn{\nabla f(x^k) - g^k} + 2\gamma L \sqn{g^k}.
    \end{align*}
    Plugging this in to the previous expression we obtain the desired result.
\end{proof}
\begin{lemma}\label{lem:bound-gradient}
    Let $f: \mathbb{R}^d \to \mathbb{R}$ be an $L$-smooth function with a lower bound \( f^\star \). Then, for any $x \in \mathbb{R}^d$,
    \[
        \|\nabla f(x)\|^2 \leq 2L(f(x) - f^\star).
    \]
\end{lemma}
\begin{proof}
    By \Cref{as:L-smooth}, for any \( y \), we have:
    \begin{align*}
        f(y) & \leq f(x) + \ev{\nabla f(x), y - x} + \frac{L}{2} \|y - x\|^2.
    \end{align*}
    We choose \( y = x - \frac{1}{L} \nabla f(x) \), and obtain
    \begin{align*}
        f\left(y\right) & \leq f(x) - \frac{1}{2L} \|\nabla f(x)\|^2.
    \end{align*}
    Since \( f(y) \geq f^\star \), we re-arrange and obtain the result.
\end{proof}

\thmCAFe* 
\begin{proof}
    Let us denote $\ec{f(x^{k+1}) + \frac{\gamma}{2\br{1-\omega}} \sqn{\overline{e}^{k+1}}} := \Psi^{k+1}$.
    Then, if we start from \Cref{eq:main-recursion}, and add the result from \Cref{lem:compression} multiplied by $\frac{\gamma}{2\br{1-\omega}}$, we have
    \begin{align}
        \Psi^{k+1} & \leq - \frac{\gamma}{2} \br{1+\frac{\omega}{1-\omega}} \ec{\sqn{\nabla f(x^k)}} \nonumber             \\
                   & - \br{\frac{1}{2\gamma} - \frac{L}{2} - \frac{L\omega}{1-\omega}} \ec{\sqn{x^{k+1} - x^k}}  \nonumber \\
                   & + \frac{\gamma}{2}\cdot\frac{\omega B^2}{1-\omega} \ec{\sqn{\nabla f(x^{k+1})}} + \Psi^k.
    \end{align}
    If $\gamma$ satisfies \Cref{eq:gamma-condition}, we can ignore the second term.
    Unrolling the recursion for $K$ iterations, we obtain
    \begin{align}
        \Psi^K & \leq \Psi^0 - \frac{\gamma}{2} \br{1+\frac{\omega}{1-\omega}} \sum_{k=0}^{K-1} \ec{\sqn{\nabla f(x^k)}} \nonumber \\
               & + \frac{\gamma}{2}\cdot\frac{\omega B^2}{1-\omega} \sum_{k=0}^{K-1} \ec{\sqn{\nabla f(x^{k+1})}}.
    \end{align}
    Simplifying, and since the compression error is null at zero,
    \begin{align}
        f(x^K) & \leq f(x^0) + \frac{\gamma}{2}\cdot\frac{\omega B^2}{1-\omega} \ec{\sqn{\nabla f(x^{K})}}  \nonumber     \\
               & - \frac{\gamma}{2} \br{1+\frac{\omega \br{1-B^2} }{1-\omega}} \sum_{k=0}^{K-1} \ec{\sqn{\nabla f(x^k)}}.
    \end{align}
    Next, if we use \Cref{lem:bound-gradient} to bound the $\ec{\sqn{\nabla f(x^{K})}}$ term, note that $\frac{\gamma \omega B^2 L}{1-\omega}  \leq 1$ is always satisfied since $\omega B^2 < 1$ and \Cref{eq:gamma-condition} imply it.
    Thus, we obtain
    \begin{align*}
        \frac{\gamma}{2} \br{1+\frac{\omega \br{1-B^2} }{1-\omega}} \sum_{k=0}^{K-1} \ec{\sqn{\nabla f(x^k)}} \leq f(x^0) - f^\star.
    \end{align*}
    Re-arranging, we obtain the desired result.
\end{proof}

\end{document}